%% file: main.tex
\documentclass[letterpaper, 10 pt, conference]{ieeeconf}  

\IEEEoverridecommandlockouts                              

\usepackage{graphics} 
    \DeclareGraphicsExtensions{.jpg,.pdf,.mps,.png} 
    \graphicspath{{fig/} {./}} 
\usepackage{amsmath} 
\usepackage{amssymb}  
\usepackage{setspace} 
\usepackage{bm} 
\usepackage{flushend} 

\input{macros.tex}  
\usepackage{siunitx} 
\usepackage{gensymb} 
\usepackage{multirow} 
\usepackage{caption}
\usepackage{subcaption}
\usepackage{colortbl} 
\usepackage{xcolor} 

\usepackage{soul}
\usepackage{framed} 
\usepackage{svg}

\usepackage{pgfplots}
\pgfplotsset{compat=newest}
\usetikzlibrary{plotmarks}
\usetikzlibrary{arrows.meta}
\usepgfplotslibrary{patchplots}
\usepackage{xcolor}

\definecolor{mygreen}{rgb}{0.0,1.0,0}
\definecolor{mylightgreen}{rgb}{0.7,0.9,0.0}
\definecolor{myyellow}{rgb}{1.0,1.0,0.2}
\definecolor{myorange}{rgb}{1.0,0.5,0}
\definecolor{myred}{rgb}{1.0,0,0}

\usepackage{grffile}
\pgfplotsset{plot coordinates/math parser=false}
\newlength\fwidth
\newlength\fheight

\usepackage{cite}

\newcommand{\auths}[1]{{#1} {$et$ $al.$}}

\title{\LARGE \bf
Online Inertia Parameter Estimation for Unknown Objects \\ Grasped by a Manipulator Towards Space Applications}

\author{Akiyoshi Uchida$^{1}$, Antonine Richard$^{2}$, Kentaro Uno$^{1}$, Miguel Olivares-Mendez$^{2}$ and Kazuya Yoshida$^{1}$
\thanks{$^{1}$A. Uchida, K. Uno, and K. Yoshida are with the Space Robotics Lab. (SRL) in the Department of Aerospace Engineering, Graduate School of Engineering, Tohoku University, Sendai 980-8579, Japan.}%
\thanks{E-mail: {\tt\small uchida.akiyoshi.s3@dc.tohoku.ac.jp}}
\thanks{$^{2}$ A. Richard, M. Olivares-Mendez are with The Space Robotics Research Group at the Interdisciplinary Research Center for Security reliability and Trust (SnT) in the University of Luxembourg.
}%
\thanks{E-mail: {\tt\small antoine.richard@uni.lu}}
\thanks{\textit{Corresponding author is Akiyoshi Uchida.}}
}

\begin{document}

\maketitle
\thispagestyle{empty}
\pagestyle{empty}


\begin{abstract}


Knowing the inertia parameters of a grasped object is crucial for dynamics-aware manipulation, especially in space robotics with free-floating bases. This work addresses the problem of estimating the inertia parameters of an unknown target object during manipulation. We apply and extend an existing online identification method by incorporating momentum conservation, enabling its use for the floating-base robots. The proposed method is validated through numerical simulations, and the estimated parameters are compared with ground-truth values. Results demonstrate accurate identification in the scenarios, highlighting the method’s applicability to on-orbit servicing and other space missions.

\end{abstract}



\section{INTRODUCTION}
\label{sec:intro}
\subsection{Background}
\label{subsec:background}

Precise estimation of inertia parameters is essential for achieving accurate and reliable manipulator control, as it allows control algorithms to predict and adapt to the system’s dynamic behavior under varying conditions. Accurate inertia estimates enhance trajectory planning, disturbance rejection, and overall performance, particularly in model-based control approaches such as model predictive control (MPC), where the control effectiveness strongly depends on the fidelity of the system’s dynamic model~\cite{MAYNE20142967,QIN2003733,Rybus2017}.



Inertia parameter identification becomes even more crucial for orbital servicing robots equipped with manipulators. Due to the microgravity environment and the robot's floating base, the motion of the manipulator influences the robot's attitude. As a result, the conventional approach used for ground robots—where the manipulator's trajectory is controlled by solving inverse kinematics through geometric calculations—is not directly applicable. The dynamics of the system in space require a more comprehensive control strategy that accounts for the coupling between the manipulator's motion and the robot's overall attitude~\cite{FLORESABAD20141}. \auths{Umetani} developed a control method for free-floating robots based on the resolved motion control concept and derived a Generalized Jacobian that accounts for momentum conservation in space~\cite{Umetani1989,Yoshida1991}. This technique enables the robot to control its arm while considering the motion effects on the base. However, it assumes perfect knowledge of the system's dynamics model, and as such, the accuracy of the method decreases if there are errors in the inertia parameters. The performance is further compromised when attempting to grasp a target with unknown inertia.

\begin{figure}[t]
  \centering
  \includegraphics[width=\linewidth]{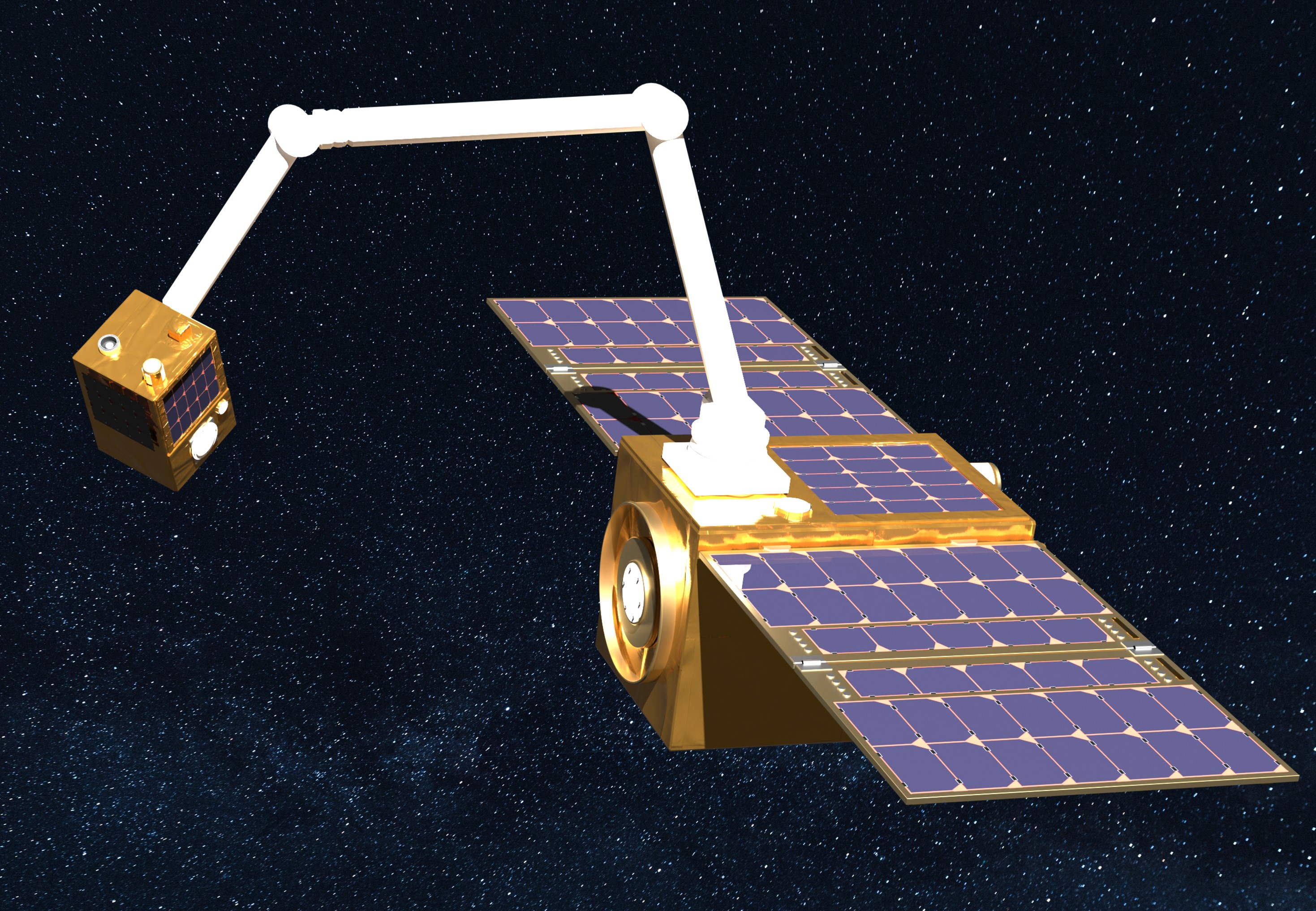}
  \caption{A conceptual illustration of a base-floating orbital robot equipped with a robot arm capturing space debris.}\label{fig:concept}
\end{figure}

{Inertia parameter identification has been a fundamental topic in robotics since its early stages~\cite{atkeson1986estimation,Pradeep1985,app11094303,wensing2017linear,Taeyoon2018,Taeyoon2020,Rucker2022,Taeyoon2018IROS,namhoon2024}. Although the relationship between forces and inertia parameters is linear, ensuring physical consistency introduces additional challenges, as the parameters must satisfy constraints such as positive definiteness and non-negative mass~\cite{app11094303,wensing2017linear}. To address this issue, geometric approaches have been proposed that model the parameter space as a Riemannian manifold, providing coordinate-invariant and physically meaningful regularization~\cite{Taeyoon2018,Taeyoon2020,Rucker2022,Taeyoon2018IROS}.
Building on this idea, \auths{Cho} introduced an online recursive formulation called RLS with log-determinant divergence regularization (RLS log-det-div), which preserves physical consistency while improving robustness in real-time inertia identification~\cite{namhoon2024}.}

The inertia estimation problem has also been studied intensively also in the space robotics field.
While certain parameters, such as the center of mass, can be estimated passively through visual observations~\cite{uno2024case}, other properties like mass and inertial tensor require physical interaction and external force application. This highlights the necessity of manipulation-based inertia estimation in space environments.
To address this challenge, various identification methods have been developed both to reduce the inertia parameter errors of the robot itself and to estimate the inertia properties of the target debris. Most of these approaches are based on the principle of momentum conservation, which fundamentally distinguishes orbital robots from their terrestrial counterparts~\cite{ma2008orbit,MENG20201792,XU2017131,FLORESABAD20141,Bingyu2023,GANGAPERSAUD20193900,Abiko2007,Lampariello2005,Nguyen2013}.
{\auths{Abiko} developed an online adaptation law integrated with momentum-based control for post-grasp stabilization~\cite{Abiko2007}. The method used Lyapunov stability theory to update the inertial parameters in real time by exploiting the momentum coupling between the base and the manipulator in a free-floating system.
\auths{Xu} proposed an on-orbit identification approach that models the system as an equivalent single-body (all joints locked) or two-body (one joint released) system, and estimates the complete inertia parameters through a nonlinear optimization solved by Particle Swarm Optimization (PSO)~\cite{XU2017131}.
These studies demonstrated the effectiveness of momentum conservation–based identification for free-floating systems, but did not explicitly ensure the physical consistency of the estimated inertia parameters.}

\subsection{Motivation}
\label{subsec:motivation}

For feasible identification of inertia parameters by formulating the problem as a least squares estimation, incorporating a physically reasonable regularizer is essential, as demonstrated in previous works on ground robots~\cite{Taeyoon2018,Taeyoon2020,Taeyoon2018IROS,Rucker2022,namhoon2024}. In contrast, most inertia parameter estimation methods for space robots rely on momentum conservation principles, a distinctive characteristic of such systems. However, regularization to ensure that the estimated inertia remains within its physically valid manifold has not been widely considered.

Therefore, this work aims to establish a feasible approach for inertia identification in space debris by incorporating the concept of constraining estimated parameters within a physically meaningful space. We begin by applying RLS with log-determinant divergence regularization~\cite{namhoon2024} for object inertia estimation using a ground-based robot and subsequently extending it to a floating robotic system (see \fig{fig:concept}). In~\cite{namhoon2024}, the method was validated by applying it to estimate the inertia of a quadruped robot leg, where the initial guess of the inertia was obtained from its CAD model. However, in space debris removal missions, such prior information is not usually accurate due to its possible physical damages and its propellant consumption, which necessitates in-situ direct estimation.

Furthermore, to extend inertia identification techniques to orbital servicing scenarios, incorporating a momentum conservation term into the objective function is beneficial. Most inertia optimization methods for ground robots rely on the relationship between joint space forces and inertia parameters. However, forces are more susceptible to disturbances, making the estimation process more sensitive. Additionally, the regressor matrix, which relates force to inertia, includes system acceleration, further contributing to estimation instability. By using momentum as a regressor instead, inertia parameters can be estimated without relying on force and acceleration, which tend to exhibit high variability.




\subsection{Contribution}
\label{subsec:contributions}
In this study, we demonstrate RLS log-det-div can be effectively applied to identify the inertia parameters of an unknown object. Furthermore, we extend its applicability to in-orbit mission scenarios based on momentum conservation, particularly in capturing space debris. The contributions of this study are highlighted as follows:

\begin{itemize}
    \item We apply RLS log-det-div to estimate the inertia parameters of a grasped object by fixed manipulators and numerically evaluate its accuracy.
    \item We demonstrate that this method can also be used for free-floating space robots to estimate the inertia parameters of uncooperative space debris.
    \item We extend the previous approach for space applications by incorporating a momentum conservation term, leading to improved robustness against disturbances and more stable inertia estimation.
\end{itemize}

\section{PRELIMINARIES}

\subsection{Dynamics}
The equation of the robot dynamics is formulated as follows:

\begin{equation}
    \bm{M}(\bm{\bm{q}})\dot{\bm{\bm{\nu}}} + \bm{c}(\bm{\bm{q}}, \bm{\bm{\nu}})\bm{\nu} + \bm{g}(\bm{q}) = \bm{F}
    \label{eq:dynamics}
\end{equation}

\noindent where $\bm{q} \in Q$ represents the generalized coordinates of the system in the configuration manifold $Q$, $\bm{\nu} \in \mathbb{R}^{n_\text{d}}$ is the generalized velocity, $\bm{M}(\bm{q}) \in \mathbb{R}^{n_\text{d} \times n_\text{d}}$ is the mass matrix, $\bm{c}(\bm{q}, \bm{\nu}) \in \mathbb{R}^{n_\text{d} \times n_\text{d}}$ represents the Coriolis and centrifugal terms, $\bm{g}(\bm{q}) \in \mathbb{R}^{n_\text{d}}$ represents the generalized gravitational forces, and $\bm{F} \in \mathbb{R}^{n_\text{d}}$ is the generalized force applied to the system externally, with $n_\text{d}$ as degrees of freedom. This equation represents the general form of a robotic system's dynamics, applicable to both terrestrial and space robots. 
{In the case of a floating-base system, the generalized coordinates include the base position and orientation, and the corresponding generalized velocity accounts for the linear and angular momentum of the base. In this work, all quantities related to the base are expressed in its local body-fixed frame, unless otherwise stated.}

It is known that the inertia parameters of the robot are linearly associated with the generalized force using a regressor matrix~\cite{LU240203}. Based on this knowledge, \eq{eq:dynamics} can be reconstructed as shown in the following equation:

\begin{equation}
    \bm{U}(\bm{q}, \bm{\nu}, \dot{\bm{\nu}}) \bm{\Phi} = \bm{F}
    \vspace{-0mm}
    \label{eq:linear-known}
\end{equation}

\noindent where, \( \bm{U}(\bm{q}, \bm{\nu}, \dot{\bm{\nu}}) \in \mathbb{R}^{n_\text{d} \times 10n_\text{b} } \) denotes the regressor matrix, with $n_\text{b}$ representing the number of the links. The vector form of inertia parameters of the robot, \( \bm{\Phi}\in\mathbb{R}^{10n_\text{b}} \), is the correction of \( \bm{\phi} \in\mathbb{R}^{10}\), which is the parameter of each \( \mathrm{link}_i \) as defined in the following equations. Please note that the link index $i\in\{0, 1, ..., n_\mathrm{b}-1\}$ starts from 0 for the base link, and ends at $n_\mathrm{b}-1$ for the end effector. $\bm{\phi}$ is defined as follows where $m \in \mathbb{R}$, $\bm{h} = m \bm{c} \in \mathbb{R}^3$ and ${\bm{I}} \in \mathbb{R}^{3 \times 3}$ are mass, a first momentum vector which is a product of mass $m$ and the position vector of the center of the mass $\bm{c} \in \mathbb{R}^3$, and inertia tensor around the origin of the link frame respectively. All parameters are defined in the coordinate frame \{$i$\} attached to the \( \mathrm{link}_i \), with the origin located at the parent joint:

\begin{align}
        \bm{\Phi} &\triangleq  [{\bm{\phi}^0}^\top {\bm{\phi}^1}^\top \cdots  {\bm{\phi}^{n_\text{b}-1}}^\top]^\top \\
    \bm{\phi}& = [m, h_x, h_y, h_z, I_{xx}, I_{yy}, I_{zz}, I_{yz}, I_{zx}, I_{xy}]^\top   
    \label{eq:defphi}
\end{align}

\noindent where,

\begin{equation}
    {\bm{I}} = 
    \begin{bmatrix}
        I_{xx} & I_{xy} & I_{xz} \\
        I_{xy} & I_{yy} & I_{yz} \\
        I_{xz} & I_{yz} & I_{zz}  
    \end{bmatrix}\begin{matrix} \\ \\.\end{matrix}
\end{equation}

Let's assume that a force of unknown magnitude in generalized coordinates $\bm{y}_\psi\in\mathbb{R}^{n_\mathrm{d}}$ can be expressed as $-\bm{\Lambda} \bm{\psi}$ where $\bm{\Lambda}\in\mathbb{R}^{n_{\mathrm{d}}\times n_\mathrm{p}}$ is the regressor matrix for the unknown parameters and $\bm{\psi} \in \mathbb{R}^{n_\mathrm{p}}$ is the vector of unknown parameters, with using $n_\mathrm{p}$ as a number of the unknown parameters such as friction coefficients. Under the condition of this assumption, the known force element $\bm{y}\in\mathbb{R}^{n_\mathrm{d}}$ is described as $\bm{y} = \bm{F} + \bm{\Lambda} \bm{\psi}$. With this expression, \eq{eq:linear-known} is written as shown in the following equation where $\bm{\Gamma}(\bm{q}, \bm{\nu}, \dot{\bm{\nu}}) \triangleq [\bm{U}(\bm{q}, \bm{\nu}, \dot{\bm{\nu}}) \; \bm{\Lambda}(\bm{q}, \bm{\nu}, \dot{\bm{\nu}})]\in\mathbb{R}^{n_{\mathrm{d}}\times (10_\mathrm{b}+n_\mathrm{p})}$ and $\bm{\theta} \triangleq [\bm{\Phi} \; \bm{\psi}]^\top\in\mathbb{R}^{10n_\mathrm{b}+n_\mathrm{p}}$:

\begin{equation}
    \bm{\Gamma} \bm{\theta} = \bm{y}
    \label{eq:regressor}.
\end{equation}

\subsection{Minimal parameter set and identifiability}

In general, the vector of inertial parameters $\bm{\Phi}$ contains redundancies due to linear dependence among the columns of the regressor. Thus, identification based on the full parameter set may suffer from poor conditioning and ambiguity. To address this, we consider the concept of the base parameter set, which consists of a minimal, linearly independent subset of parameters that are sufficient to describe the system dynamics~\cite{gautier1988}. 
{Furthermore, the identifiability of the inertia parameters depends on the richness of the excitation in the motion. When the robot’s trajectories are not sufficiently diverse, the regressor matrix loses rank, and some parameters cannot be uniquely determined. The mass ratio between the manipulator and the grasped object also influences sensitivity; if the object is much lighter than the arm, its effect on the overall dynamics becomes negligible, leading to poor conditioning. In addition, measurement noise and modeling errors, such as unmodeled joint friction or structural compliance, can distort the regression relationship and further reduce identifiability.}


\subsection{Recursive least squares with log-determinant divergence regularization}
{The key idea of RLS log-det-div is to recursively update the parameter estimates by minimizing the squared prediction error, while simultaneously penalizing deviations in the estimated inertia matrix using the log-determinant divergence. This divergence quantifies the geometric distance between positive definite matrices, thereby ensuring that the updated inertia parameters remain within the space of physically realizable matrices and enhancing robustness against noisy or incomplete measurements~\cite{namhoon2024}.}

It is known that the scalar inequalities that physically consistent inertia parameter $\bm{\phi}$ needs to satisfy can be expressed as $\bm{L}(\bm{\phi}) > 0$~\cite{wensing2017linear}, where pseudo-inertia matrix $\bm{L}(\bm{\phi}) \in \mathbb{R}^{4\times 4}$ is described as:

\begin{equation}
    \bm{L}(\bm{\phi}) \triangleq 
    \begin{bmatrix}
        \bm{\Sigma} & \bm{h} \\
        \bm{h}^\top    & \bm{\Sigma}
    \end{bmatrix}
\end{equation}

\noindent with defining $\bm{\Sigma}$ as:

\begin{equation}
    \bm{\Sigma} \triangleq \frac{1}{2} \mathrm{tr}({\bm{I}}) \mathbb{I}_3 - {\bm{I}}
\end{equation}

\noindent where $\mathbb{I}_n \in \mathbb{R}^{n \times n}$ denotes the $n$-dimensional identity matrix.

Given the regularized least squares objective function $J_N(\theta)$ as expressed in \eq{eq:basic_objective}, the inertia identification algorithm using RLS with log-determinant divergence regularization is driven by solving the $\bm{\Delta}_k\in\mathbb{R}^{10n_\mathrm{b}+n_\mathrm{p}}$, which minimizes $J_k(\bm{\theta}_k + \bm{\Delta}_k)$ at time step $k$:

\begin{flalign}
    J_N(\bm{\theta}) &=\frac{1}{2} \sum^N_{j=1} (\bm{y}_j - \bm{\Gamma}_j\bm{\theta})^\top \bm{W}_j (\bm{y}_j - \bm{\Gamma}_j\bm{\theta}) \nonumber \\
                &\quad -\frac{1}{2} \sum^N_{j=1} (\bm{\theta} - \hat{\bm{\theta}}_j)^\top \bm{G}_j (\bm{\theta} - \hat{\bm{\theta}}_j)
                +\alpha_N R(\bm{\theta})
    \label{eq:basic_objective}
\end{flalign} 

\noindent where $N$ is the number of the sampled measurements; $\bm{y}_j \triangleq \bm{y}(t_j)$ is the output; $\bm{\Gamma}_j\triangleq\bm{\Gamma}(\bm{q}(t_j), \bm{\nu}(t_j), \dot{\bm{\nu}}(t_j))$ is the regressor matrix; $\bm{\hat{\theta}}_j\triangleq\bm{\hat{\theta}}(t_j)$ denotes the estimated parameters; $\bm{W}_j = \bm{W}_j^\top > 0$ is the weight matrix; $\bm{G}_j = \bm{G}_j^\top > 0$ is the forgetting factor; $\alpha_N$ is the regularization strength at step $N$; and $R(\bm\theta)$ is a twice differentiable convex regularizer at the $j$-th time step.

The regulariser $R(\bm{\theta})$ is defined as expressed in \eq{eq:regulariser}, using a log-determinant divergence function as described in \eq{eq:logdetdiv}. This function represents the distance of $\bm{\phi}^i$ and its prior value $\bm{\hat{\phi}}^i_0$ in the manifold. The use of log-det divergence as a regulariser allows estimation to stay in physically consistent space contributing to stable estimation:

\begin{equation}
    R(\bm{\theta}) = \sum^{n_\text{b}-1}_{i=0}D_\sigma ( \bm{L}(\bm{\phi}^i), \bm{L}(\hat{\bm{\phi}^i_0}))
              + \frac{\beta}{2} \vert\vert \bm{\psi}-\hat{\bm{\psi}_0} \vert\vert
    \label{eq:regulariser}
\end{equation}

\begin{flalign}
    &D_\sigma(\bm{L}(\bm{\phi}^i), \bm{L}(\hat{\bm{\phi}^i_0}))  \nonumber \\
    &= -\mathrm{log}\frac{\vert \bm{L}(\bm{\phi}^i) \vert}{\vert \bm{L}(\hat{\bm{\phi}^i_0}) \vert}
    + \mathrm{tr}(\bm{L}(\hat{\bm{\phi}^i_0})^{-1}\bm{L}(\bm{\phi}^i)) - 4
    \label{eq:logdetdiv}.
\end{flalign}

By solving $\bm{\Delta}_k$ at each time step using Newton-Raphson iterations, the estimated parameters $\hat{\bm{\theta}_k}$ can be obtained at each time by computing $\hat{\bm{\theta}}_k = \hat{\bm{\theta}}_{k-1} + \bm{\Delta}_k$. 



\section{MOMENTUM CONSERVATION-BASED REGRESSION}
In orbital servicing, the momentum of the system is conserved without the use of thrusters, and it can be used as a regressor. Momentum-based regression offers two main advantages: robustness to force disturbances, and independence from acceleration measurements. The momentum of the robot around the base $\bm{P}_\text{b} \in \mathbb{R}^6$ can be expressed as shown in the following equation, where $\bm{M}_{(1:6,:)}(\bm{q}) \in \mathbb{R}^{6 \times n_\text{d}}$ represents the top 6 rows of the inertia matrix $\bm{M}(\bm{q})$:

\begin{equation}
    \bm{M}_{(1:6,:)}(\bm{q}) {\bm{\nu}} = \bm{P}_\text{b}.
\end{equation}

The mass matrix of the floating robot can be expressed as follows:

\begin{equation}
    \bm{M}(\bm{q}) = \bm{\mathcal{J}}(\bm{q})^\top \bm{\mathcal{G}} \bm{\mathcal{J}}(\bm{q})
\end{equation}

\noindent where $\bm{\mathcal{J}} \in \mathbb{R}^{6n_\mathrm{b}\times n_\mathrm{d}}$ is a collection of the Jacobian $\bm{\mathcal{J}}_i \in \mathbb{R}^{6\times n_\mathrm{d}}$, and $\bm{\mathcal{G}} \in \mathbb{R}^{6n_\mathrm{b}\times 6n_\mathrm{b}}$ is the collection of the spacial inertia matrix $\bm{\mathcal{G}}_i \in \mathbb{R}^{6\times6}$ for each $\mathrm{link}_i$ as shown in the following equations:

\begin{align}
    \bm{\mathcal{J}}(\bm{q}) =& \begin{bmatrix}
        \bm{\mathcal{J}_0}(\bm{q})^\top & \bm{\mathcal{J}_1}(\bm{q})^\top & \cdots & \bm{\mathcal{J}}_{n_\mathrm{b}-1}(\bm{q})^\top
    \end{bmatrix}^\top \\
    \bm{\mathcal{G}} =& \begin{bmatrix}
        \bm{\mathcal{G}_0}  &  \bm{0}_{6\times6}  & \cdots & \bm{0}_{6\times6} \\
        \bm{0}_{6\times6}   &  \bm{\mathcal{G}}_1 & \cdots & \bm{0}_{6\times6} \\
        \vdots & \vdots & \ddots & \vdots \\
        \bm{0}_{6\times6} & \bm{0}_{6\times6} & \cdots & \bm{\mathcal{G}}_{n_\mathrm{b}-1}
    \end{bmatrix}\begin{matrix} \\ \\ \\ .\end{matrix}
\end{align}


The Jacobian matrix $\bm{\mathcal{J}}(\bm{q})$ depends solely on the robot’s kinematic structure and its generalized coordinates, and is therefore independent of the inertia parameters. Furthermore, since the spatial inertia matrix $\bm{\mathcal{G}}$ depends linearly on the inertia parameter vector $\bm{\Phi}$, it is also possible to derive a regressor matrix that linearly relates the momentum around the base to the inertia parameters as follows, similar to the formulation discussed in the previous section:

\begin{equation}
    \bm{U}_\text{m}(\bm{q}, \bm{\nu}) \bm{\Phi} = \bm{P}_b
    \label{eq:momenta_regressor}
\end{equation}

\noindent where $\bm{U}_\text{m}(\bm{q}, \bm{\nu}) \in \mathbb{R}^{6 \times 10n_\text{b}}$ is the regressor matrix for the momentum. 

Since the $k$-th column of the regressor, $\bm{U}_{\text{m}(:,k)}$, is linearly associated with the $k$-th element of the vector of inertia parameters $\bm{\Phi}_{(k)}$, it can be expressed as:

\begin{equation}
   \bm{U}_{\text{m}(:,k)}(\bm{q}, \bm{\nu}) = \bm{\mathcal{J}}_{(:,1:6)}(\bm{q})^\top \frac{\partial \bm{\mathcal{G}}}{\partial \bm{\Phi}_{(k)}}  \bm{\mathcal{J}}(\bm{q}) \bm{\nu}
\end{equation}

\noindent where $\bm{\mathcal{J}}_{(:,1:6)}(\bm{q})$ is the left 6 columns of the Jacobian $\bm{\mathcal{J}}(\bm{q})$.

With this matrix, the objective function formulated in \eq{eq:basic_objective} can be rewritten as expressed in the following equation, under the condition that the initial momentum around the base is zero:

\begin{flalign}
    &J_{\mathrm{m}N}(\bm{\Phi}) =\frac{1}{2} \sum^N_{j=1} \bm{\Phi}^\top{\bm{U}_\text{m}}_j^\top \bm{W}_{\mathrm{m}j} {\bm{U}_\text{m}}_j\bm{\Phi} \nonumber \\
                &\quad -\frac{1}{2} \sum^N_{j=1} (\bm{\Phi} - \hat{\bm{\Phi}}_j)^\top \bm{G}_{\mathrm{m}j} (\bm{\Phi} - \hat{\bm{\Phi}}_j)
                +\alpha_{\mathrm{m}N} R_{\mathrm{m}}(\bm{\Phi})
    \label{eq:momentum_objective}
\end{flalign}

\noindent where $J_{\mathrm{m}N}(\bm{\Phi})$ is the objective function; $\bm{U}_{\mathrm{m}j}\triangleq\bm{U}_\mathrm{m}(\bm{q}(t_j), \bm{\nu}(t_j))$ is the regressor matrix; $\bm{\hat{\Phi}}_j\triangleq\bm{\hat{\Phi}}(t_j)$ denotes the estimated inertia parameters; $\bm{W}_{\mathrm{m}j} = \bm{W}_{\mathrm{m}j}^\top > 0$ is the weight matrix; $\bm{G}_{\mathrm{m}j} = \bm{G}_{\mathrm{m}j}^\top > 0$ is the forgetting factor; $\alpha_{\mathrm{m}N}$ is the regularization strength at step $N$; and $R(\bm\Phi)$ is a twice differentiable convex regularizer at the $j$-th time step, in case of momentum regression. 
{It is worth noting that the proposed approach remains applicable even with non-zero initial momentum, since taking momentum differences cancels the constant offset, or initial momentum, while preserving the regression structure.}
Here, $R_\mathrm{m}(\bm{\Phi})$ is defined in the same way as \eq{eq:regulariser} as follows, with the unknown parameter term zero:

\begin{equation}
    R_\mathrm{m}(\bm{\Phi}) = \sum^{n_\text{b}-1}_{i=0}D_\sigma ( \bm{L}(\bm{\phi}^i), \bm{L}(\hat{\bm{\phi}^i_0}))
    \label{eq:regulariser_momentum}.
\end{equation}


By using the momentum-based regressor matrix $\bm{U}_\mathrm{m}(\bm{q}, \bm{\nu})$ as formulated in \eq{eq:momenta_regressor}, instead of the force-based regressor $\bm{\Gamma}(\bm{q}, \bm{\nu}, \bm{\dot{\nu})}$ in \eq{eq:regressor}, the estimation process can be performed without relying on acceleration measurements. Since $\bm{\Gamma}(\bm{q}, \bm{\nu}, \bm{\dot{\nu})}$ explicitly depends on system accelerations, which are typically sensitive to sensor noise and external disturbances, conventional force-based methods inherently suffer from estimation instability. In contrast, the momentum-based approach utilizes only the robot’s configuration and velocity, making the regression matrix $\bm{U}_\mathrm{m}(\bm{q}, \bm{\nu})$ inherently robust against such disturbances.

Furthermore, it is important to note that, in conventional methods, the least squares objective function $J_N(\bm{\theta})$ in \eq{eq:basic_objective} is formulated with an extended parameter vector $\bm{\theta} = [\bm{\Phi}^\top\; \bm{\psi}^\top]^\top$, which includes not only the inertia parameters $\bm{\Phi}$ but also other unknowns such as friction coefficients $\bm{\psi}$. These additional parameters often complicate the identification process and reduce robustness due to model uncertainties. In contrast, the proposed momentum-based formulation simplifies the problem by focusing solely on the physically meaningful inertia parameters $\bm{\Phi}$, as shown in the momentum-based objective function $J_{\mathrm{m}N}(\bm{\Phi})$ in \eq{eq:momentum_objective}. This reduced dependency naturally eliminates the influence of hard-to-model parameters like joint friction and actuator dynamics, which are common sources of estimation degradation in force-based identification frameworks.

\section{NUMERICAL EXPERIMENTS}

\subsection{Experimental setup}

\subsubsection{Problem formulation}
\begin{figure}[t]
  \centering
  \includegraphics[width=\linewidth]{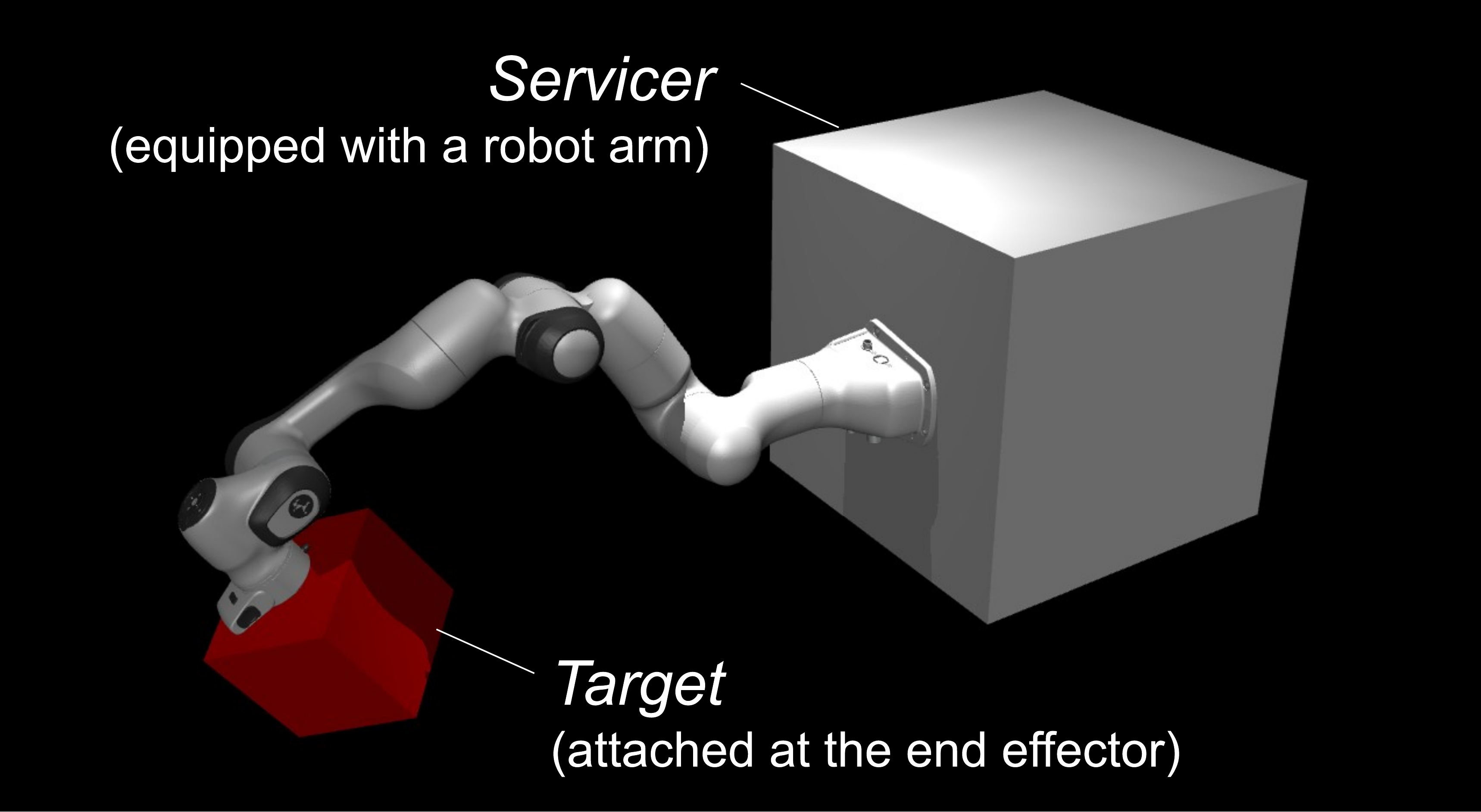}
  \caption{MuJoCo dynamics simulation to estimate unknown inertial parameters of the captured target by a base-floating orbital servicer.}\label{fig:mujoco_img}
\end{figure}


Although the method we employed was originally developed to identify the full dynamics of the system, including the robot itself, we focus on estimating 10 parameters corresponding to the inertia of the target object. These 10 parameters were selected from the 47 base parameters obtained through numerical analysis of the regressor matrices. Symbolic analysis confirmed that all 10 parameters are linearly independent, indicating that they are theoretically identifiable. We applied the RLS process exclusively to the target, which is rigidly attached to the end-effector. This approach allows us to specifically evaluate the accuracy of inertia estimation for the unknown target. 
{The proposed method assumes a rigid manipulator structure. In practice, however, modeling uncertainties can arise from various factors, such as structural flexibility, link compliance, or joint elasticity, which may affect the accuracy of inertia parameter identification. Although such flexibility is not explicitly modeled in this study, we examined the robustness of the proposed method to general model errors by introducing different levels of perturbation in the robot’s own inertia parameters during validation.} Numerical experiments were conducted in the MuJoCo~\cite{MuJoCo} simulation environment using our custom C++ implementation, based on prior work implemented in Julia~\cite{EntDivRLS}. \fig{fig:mujoco_img} shows a rendered image of the orbital robot in MuJoCo, where the red object represents the grasped target.

We formulated the problem as follows:


\begin{itemize}
    \item A target object is rigidly connected to the end-effector at the beginning of the estimation.
    {\item The manipulator is assumed to be rigid, with structural flexibility neglected.}
    \item The estimation sequence is conducted from a stable initial state of the system, assuming that the space debris has already been grasped following rendezvous and detumbling.
    \item In the case of the force-based regression, torque sensors are assumed to be available at each joint. The measurement noise is simulated by adding 5\% of unit-magnitude random noise to the ground-truth torque values.
    \item For object manipulation, which is beyond the scope of this paper, the exact inertia parameters can be used to solve the inverse dynamics.
    \item Friction and system noise are ignored for the controller, and control is assumed to perfectly follow the desired trajectory to isolate the estimation performance from controller-related effects.
    \item In the case of orbital robots, the base is free-floating with no external forces applied, and the total momentum of the system is conserved.
\end{itemize}

\subsubsection{Inertia parameter randomization}
\label{subsubsec:ramdom}
The estimation accuracy is directly influenced by the precision of the robot’s inertia parameters. To evaluate the impact of inaccurate inertia, we introduced randomness into the robot parameters as follows. The asterisk (\(*\)) in the upper right corner denotes a randomized value, and \(\epsilon\) represents a constant that determines the degree of randomization:

\begin{align}
    m^* &= m + m\epsilon_m \\
    \bm{c}^* &= \bm{c} + \bm{\epsilon}_c \\
    \bm{\lambda}_c^* &= \bm{\lambda}_c + \bm{\lambda}_c \circ \bm{\epsilon}_\lambda \\ 
    \bm{R}_c^* &= \bm{R}_c (\delta \bm{R} (\epsilon_{\gamma}, \bm{\epsilon}_{\hat{a}}))
\end{align}

\noindent where ($\circ$) means the element-wise product of the vectors.

 In the equations above, we define $\bm{\lambda}_c \in \mathbb{R}^3$ as the eigenvalues of the inertia tensor expressed at the center of mass, which is a principal moment of inertia, and $\bm{R}_c \in \text{SO}(3)$ as the rotation matrix which satisfies \eq{eq:eigen_inertia}, where $\bm{I}_c\in\mathbb{R}^{3\times3}$ represents the inertia tensor around the center of mass (CoM). Furthermore, \(\delta \bm{R}(\gamma, \bm{\hat{a}})\in \text{SO}(3)\) denotes the rotation matrix representing a rotation by an angle \(\gamma \in \mathbb{R}\) around the unit vector of axis \( \bm{\hat{a}} \in \mathbb{R}^3 \):

\begin{align}
    \bm{I}_c = \bm{R}_c^\top (\mathrm{diag}({\bm{\lambda}}_c)) \bm{R}_c     
    \label{eq:eigen_inertia}.
\end{align}

Based on the Parallel Axis Theorem, randomized inertia around the joint can be expressed as follows:

\begin{equation}
    \bm{I}^* = \bm{R}_c^{*T} (\mathrm{diag}({\bm{\lambda}}^*_c)) \bm{R}_c^* + m^*(\bm{c}^{*T}\bm{c}^*\mathbb{I}_{3} - \bm{c}^*\bm{c}^{*T}).
\end{equation}

From these randomized values, we reconstruct the randomized inertia parameter vector $\bm{\phi}^*$ as:

\begin{equation}
    \bm{\phi}^* = [m^*, m^*c^{*T}, I^*_{xx}, I^*_{yy}, I^*_{zz}, I^*_{yz}, I^*_{zx}, I^*_{xy}]^\top.
\end{equation}

Each randomization constant \( \epsilon \) is defined as follows, where \( \eta \sim \mathcal{U}(-1, 1) \) and $\bm{\eta} \sim \mathcal{U}(-1, 1)^3$ represents a random variable uniformly distributed in the interval \( \mathcal{U}(-1, 1) \), and \( \alpha_\eta \in [0, 1] \) denotes the randomization ratio. In the case of calculating $\bm{\epsilon}_\lambda$, to satisfy the triangle inequality of the principal moment of inertia, \( \mathcal{U}\) is repeatedly updated until it satisfies the condition:

\begin{equation}
    \begin{cases}
        \epsilon_m = \alpha_\eta \eta \in \mathbb{R}\\
        \bm{\epsilon}_c = 0.01\cdot m \alpha_\eta \bm{\eta} \in \mathbb{R}^3 \\
        \bm{\epsilon}_\lambda = \alpha_\eta \bm{\eta} \in \mathbb{R}^3, \quad \text{where} \quad {\lambda}_i^* + {\lambda}_j^* > {\lambda}_k^* \\
        \epsilon_\gamma = \alpha_\eta \pi \eta \in \mathbb{R}\\
        \bm{\epsilon}_{\hat{a}} = \alpha_\eta \bm{\eta}~ /~ ||\alpha_\eta \bm{\eta}|| \in \mathbb{R}^3 .
    \end{cases}
\end{equation}

\subsubsection{Object manipulation}
RLS log-det-div estimates inertia parameters based on the object's motion. In this work, we control the target object within a workspace fixed to the robot's base frame. The robot is actuated through joint velocity control, with inverse kinematics numerically solved using the Jacobian. The desired torque is then computed by solving the inverse dynamics. To have a variety of the configuration, which is essential for accurate estimation, we design the velocity $^{\mathrm{b}}\bm{\mathcal{V}}_\mathrm{t}(t) \in \mathbb{R}^6$ as follows, where 
$r =$\SI{0.15}{m}, $r_{\omega} =$\SI{0.5}{rad}, $f=$\SI{0.5}{Hz}, $f_z=$\SI{0.125}{Hz}, $f_{\omega_x}=$\SI{0.75}{Hz}, $f_{\omega_y}=$\SI{1}{Hz}, $f_{\omega_z}=$\SI{1.75}{Hz}. 
{Please note that $^{\mathrm{b}}\bm{v}_\mathrm{t}(t), ^{\mathrm{b}}\bm{\omega}_\mathrm{t}(t) \in \mathbb{R}^3$ are the linear and angular velocities of the target, respectively, expressed in the frame attached to the base of the robot rather than the target}. These parameters were selected heuristically to increase the excitation of the inertia parameters while keeping the object within the workspace of the manipulator:

\begin{equation}
    ^{\mathrm{b}}\bm{\mathcal{V}}_\mathrm{t}(t) =
\begin{bmatrix}
    ^{\mathrm{b}}\bm{v}_\mathrm{t}(t) \\ ^{\mathrm{b}}\bm{\omega}_\mathrm{t}(t)
\end{bmatrix} =
    \begin{bmatrix}
    r \cdot 2\pi f \cos( 2\pi f t) \\
    r \cdot 2\pi f \sin(2\pi f t) \\
    r \cdot 2\pi f_z \sin(2\pi f_z t) \\
    r_{\omega} \cdot 2\pi f_{\omega_x} \sin(2\pi f_{\omega_x} t) \\
    r_{\omega} \cdot 2\pi f_{\omega_y} \sin(2\pi f_{\omega_y} t) \\
    r_{\omega} \cdot 2\pi f_{\omega_z} \sin(2\pi f_{\omega_z} t) \\
    \end{bmatrix}\begin{matrix}\\ \\ \\ \\ \\ . \end{matrix}
\end{equation}

\subsubsection{Experiment cases}
We evaluated the method using both a robot fixed to the ground and an orbital robot with a free-floating base. 
\tab{tab:exp_cases} summarizes the experimental cases considered in this work, with the forgetting factor $\bm{G}_j$ set to zero. The estimation parameters were manually selected to prevent divergence and ensure a reasonable convergence speed.
For manipulation, we experimented with the Franka Emika Panda arm to interact with a target of mass \SI{5}{kg}, \SI{10}{kg}, and \SI{50}{kg}. All targets have their CoM at the geometric center and an equal mass distribution. They are represented as a cubic object with a side length of \(0.1\,\text{m}\).
 To assess the impact of inertia errors in the robot itself, we tested randomization levels of $\alpha_\eta = 0\%, 2.5\%$, as discussed in \ref{subsubsec:ramdom}. 
{For the ground experiments, we employed the conventional force-based regression method to evaluate whether the conventional approach can estimate the target mass accurately in the presence of errors in the robot’s own parameters, since momentum conservation does not hold for fixed-base robots.
Furthermore, only the \SI{5}{kg} and \SI{10}{kg} targets were used to allow stable manipulation under Earth’s gravity of \SI{9.8}{m/s^2}.
For the orbital cases, we compared the results of two approaches in the presence of errors in the robot’s inertia parameters: one using force as the regressor and the other using momentum as the regressor.}
 
The experiments for orbital robots were carried out by manipulators attached to the floating base, whose mass is \SI{216}{kg} and inertia is given by a diagonal matrix:  
\[
\bm{I}_c = \operatorname{diag}(\SI{12.96}{kg\,m^2}, \SI{12.96}{kg\,m^2}, \SI{12.96}{kg\,m^2}).
\]

We conducted experiments with a simulation and control timestep of \( \mathrm{d}t_{\text{}} = \SI{0.0001}{s} \), and an estimation timestep of \( \mathrm{d}t_{\text{est}} = \SI{0.01}{s} \).

\begin{table}[b]
    \centering    
    \caption{EXPERIMENT CASES}
    \scalebox{.72}{
    \begin{tabular}{ccccccc}
    \hline
    Robot & Scenario  & Target mass [\SI{}{kg}] & Inertia error & Regressor & $\alpha$ & $W_J$\\
    \hline
    \hline
    Panda & Ground &5 & 0.0\% & Force& 50 & $\mathbb{I}_{n_\text{d}}$ \\
    Panda & Ground &5 & 2.5\% & Force & 100  & $\mathbb{I}_{n_\text{d}}$ \\ 
    Panda & Ground &10 & 0.0\% & Force & 50  & $\mathbb{I}_{n_\text{d}}$ \\ 
    Panda & Ground &10 & 2.5\% & Force & 100  & $\mathbb{I}_{n_\text{d}}$ \\ 
    Panda & Orbital &5 & 2.5\% & Force &  100 & $\mathbb{I}_{n_\text{d}}$ \\
    Panda & Orbital &10 & 2.5\% & Force &  100 & $\mathbb{I}_{n_\text{d}}$ \\    
    Panda & Orbital &50 & 0.0\% & Force &  100 & $\mathbb{I}_{n_\text{d}}$ \\
    Panda & Orbital &50 & 2.5\% & Force &  100 & $\mathbb{I}_{n_\text{d}}$ \\    
    Panda & Orbital &5 & 2.5\% & Moment & 0.1 &  $\mathbb{I}_6$ \\
    Panda & Orbital &10 & 2.5\% & Moment & 0.1 &  $\mathbb{I}_6$ \\
    Panda & Orbital &50 & 0.0\% & Moment & 0.1 &  $\mathbb{I}_6$ \\
    Panda & Orbital &50 & 2.5\% & Moment& 0.1  & $\mathbb{I}_6$ \\
    \hline
    \end{tabular}
    }
    \label{tab:exp_cases}
\end{table}

\subsection{Results}

\subsubsection{Ground robots}

\begin{figure}[t]
    \centering
    \includegraphics[width=\linewidth]{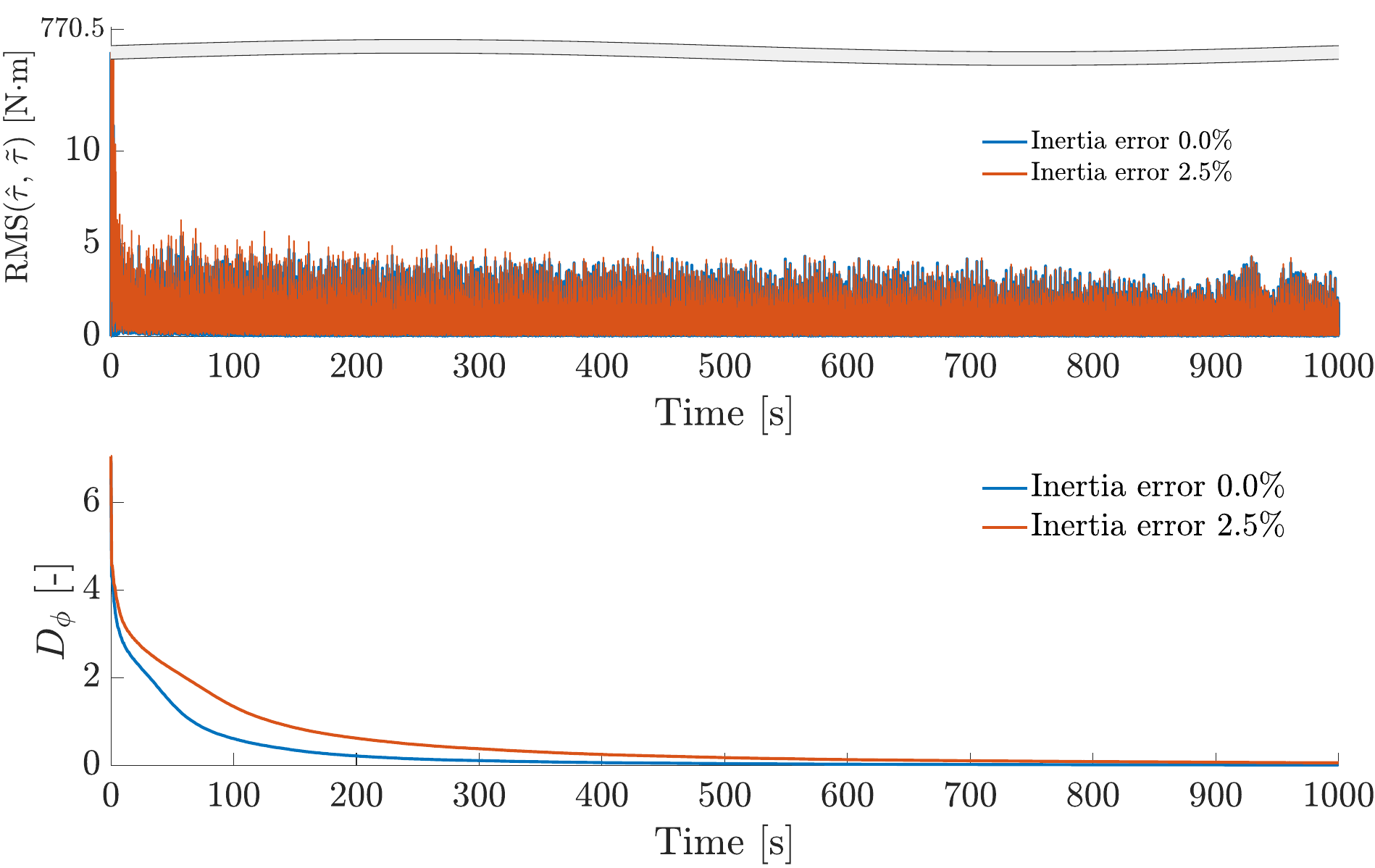}
    \caption{Parameter and force prediction errors (Fixed base, Target \SI{10}{kg}).}
    \label{fig:panda_ground_10kg}
\end{figure}

\begin{figure}
    \centering
    \includegraphics[width=\linewidth]{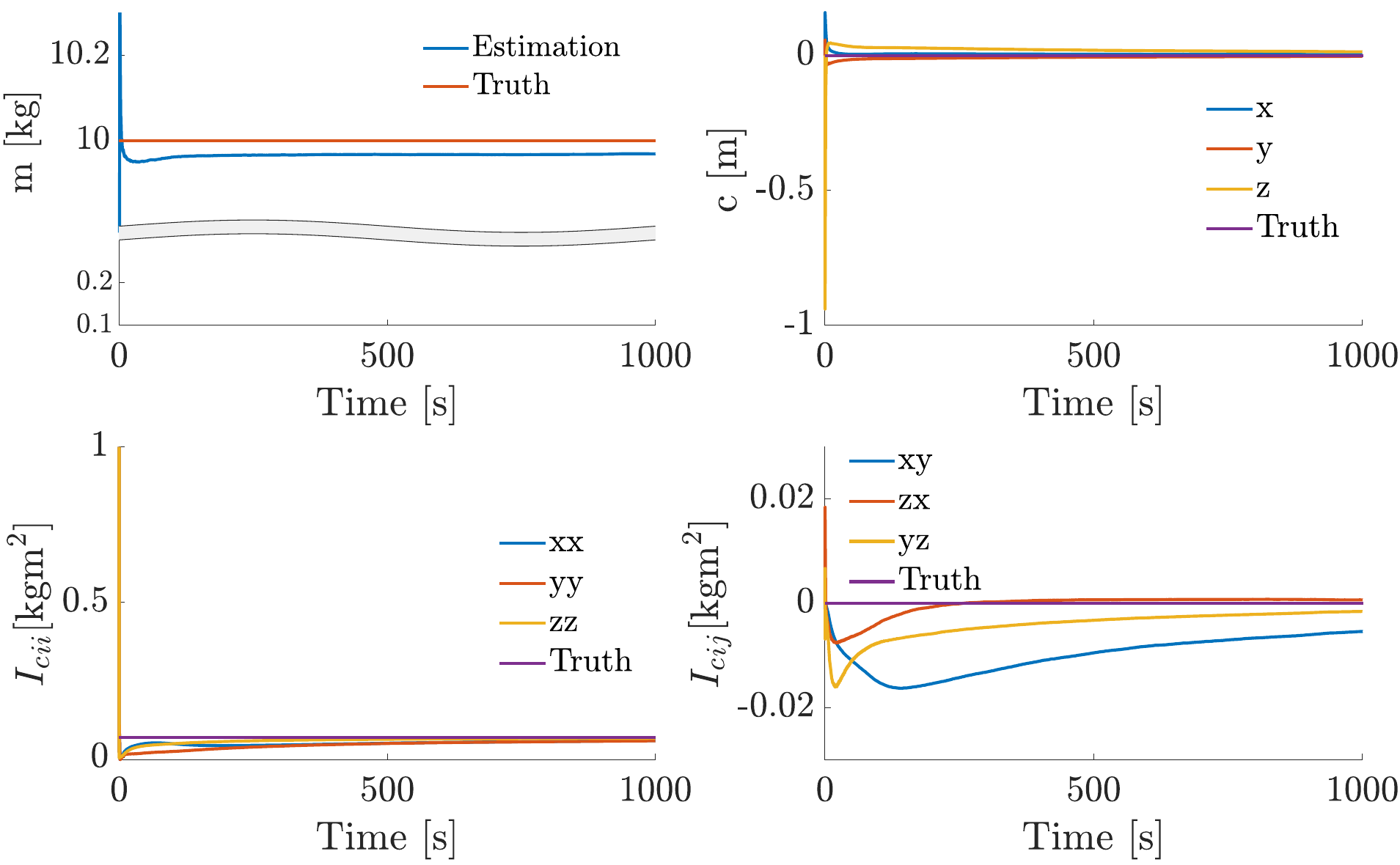}
    \caption{Comparison of predicted and actual inertia elements (Fixed base, Target 10\;kg and $\alpha_\eta=2.5\%$).}
    \label{fig:panda_ground_10kg_inertia}
\end{figure}

\fig{fig:panda_ground_10kg} shows the convergence of the estimation, presenting the errors in the predicted parameters and the estimated force in the case of a \SI{10}{kg} target grasped by the robot with its base fixed to the ground. The plot compares the results between the cases where $\alpha_\eta = 0.0\%$ and $2.5\%$. RMS($\hat{\tau}$, $\tilde{\tau}$) represents the Root Mean Square of the estimated and actual generalized force $\hat{\tau}$, $\tilde{\tau}$, while $D_\phi = D_\sigma(\bm{L}(\hat{\bm{\phi}}), \bm{L}(\tilde{\bm{\phi}}))$ is the log-determinant divergence of the real and predicted inertia parameters of the end-effector with the target. $\hat{\bm{\phi}}$ and $\tilde{\bm{\phi}}$ are the real and estimated inertia parameters of the target. Additionally, \fig{fig:panda_ground_10kg_inertia} compares each element of the target's inertia parameters around its CoM with the ground truth, in the case where $\alpha_\eta = 2.5\%$. Since the inertia is expressed relatively to the CoM, the position of the CoM should converge to zero.  
The diagonal elements of the inertia tensor, which correspond to its principal moments of inertia, are denoted as \({I_c}_{ii}\), while the off-diagonal elements are represented as \({I_c}_{ij}\).  

{\fig{fig:panda_ground_10kg} shows that the estimated inertia parameters converged to their ground-truth values, as did the estimated joint torques, even in the presence of inertia errors in the robot model. By comparing the results with and without these errors, it can be observed that a 2.5\% parameter error has only a minor impact on the estimation accuracy.
Additionally, \fig{fig:panda_ground_10kg_inertia} illustrates that each parameter converged appropriately, although the off-diagonal elements of the inertia tensor were estimated with slightly lower accuracy, which is generally more difficult to identify.}


{\tab{tab:ground_logdet_hand} presents the final values of the log-determinant divergence between the actual and estimated inertia parameters of the target. It can be seen that, without robot inertia errors, the \SI{10}{kg} target was estimated more accurately, while the \SI{5}{kg} target showed better convergence when $\alpha_\eta = 2.5\%$. This opposite trend may be attributed to the interaction between the robot’s parameter perturbations and the target-induced dynamics, which can partially cancel or amplify the effective excitation of the regressor, depending on the perturbation structure.}

\begin{table}[b]
    \centering
    \caption{Final value of the log-det divergence (Ground).}
    \scalebox{0.9}{
    \input{table/ground_logdet_hand}
    }
    \label{tab:ground_logdet_hand}
\end{table}

\subsubsection{Orbital robots}
\begin{figure}
    \centering
    \includegraphics[width=\linewidth]{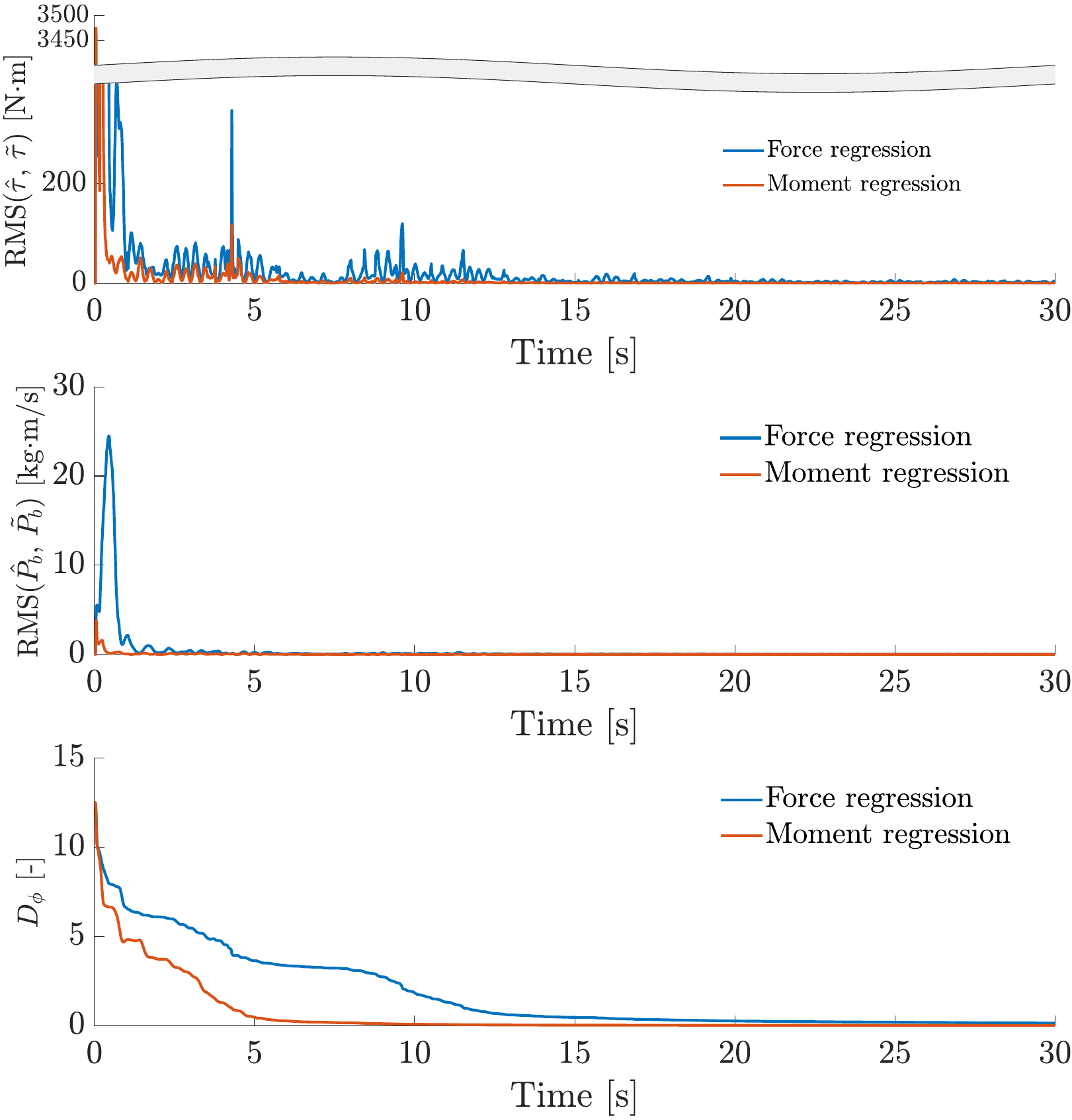}
    \caption{Parameter and force-momentum prediction errors (Floating base, Target \SI{50}{kg} and $\alpha_\eta=2.5\%$).}
    \label{fig:panda_orbital_50kg}
\end{figure}

\begin{figure}
    \centering
    \includegraphics[width=\linewidth]{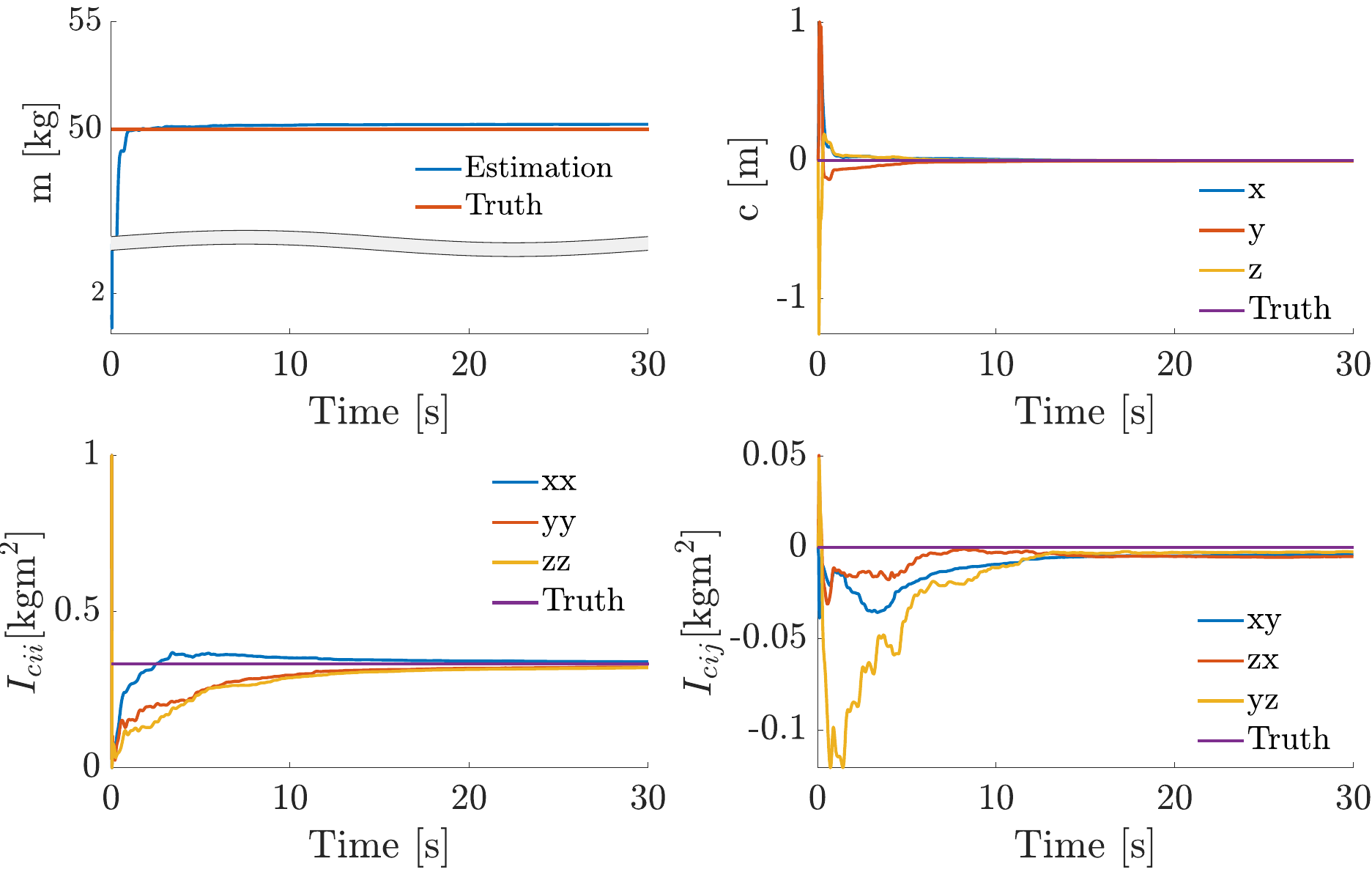}
    \caption{Comparison of predicted and actual inertia elements (Floating base, Target 50\;kg, Momentum-regression).}
    \label{fig:panda_orbital_50kg_inertia_momentum}
\end{figure}

\fig{fig:panda_orbital_50kg} and \fig{fig:panda_orbital_50kg_inertia_momentum} show the convergence of the estimation process for orbital robots in the case where the target mass is \SI{50}{kg} and the robot inertia parameters include an error of $\alpha_\eta = 2.5\%$. \fig{fig:panda_orbital_50kg} compares the prediction error of force, momentum, and log-det-div between the conventional force-based regression and the momentum-based regression proposed in this work. 
The results indicate that the momentum-based approach converges faster than the conventional force-based method. 
\fig{fig:panda_orbital_50kg_inertia_momentum} illustrates the convergence of each element of the target's inertia. All 10 parameters smoothly converged to their actual values, in contrast to the case of the ground robot. 
{This difference is particularly evident for the off-diagonal elements of the inertia tensor, which were estimated with higher accuracy in the orbital setting. Notably, the estimation in the orbital case converged approximately thirty times faster than that of the ground robot, which can be attributed to the additional degrees of freedom of the free-floating base that enrich the regressor information.}

\tab{tab:orbital_logdet_hand} presents the final values of the log-determinant divergence between the actual and predicted inertia parameters of the target.  
{The results indicate that the momentum-based regression achieved better convergence, and that the heavier target was estimated more accurately. This tendency can be explained by the fact that a heavier target induces larger dynamic reactions, resulting in a higher signal-to-noise ratio and better excitation of the regressor matrix. Consequently, the target inertia becomes more distinguishable from modeling errors in the manipulator.}

These results suggest that momentum-based regression is a robust approach, making it well-suited for orbital robots. 
{This advantage is fundamentally tied to the momentum conservation property inherent to free-floating systems in space. In the absence of external forces, the total system momentum remains constant, providing a physically consistent and low-noise dynamic relationship between the motion and the inertia parameters. In contrast, the force-based formulation for ground robots is more susceptible to external disturbances and unmodeled actuator dynamics, leading to reduced estimation stability.}

\begin{table}[b]
    \centering
    \caption{Final value of the log-det divergence (Orbital).}
    \scalebox{.9}{
    \input{table/orbital_logdet_hand}
    \label{tab:orbital_logdet_hand}
    }
\end{table}

\section{CONCLUSION}
\label{sec:conclusion}
In this work, we applied an existing online inertia estimation method to identify the completely unknown inertia of a target. We demonstrated that the method successfully identifies the target inertia for both ground and orbital robots in the presence of errors in the robot's own inertia parameters. Furthermore, we extended the method specifically for orbital applications by incorporating momentum conservation principles. 
{The proposed approach resulted in a more accurate and robust estimation compared to the existing force-based formulation that uses force as the regressor. Compared to previous studies on space robotics, such as \auths{Abiko}~\cite{Abiko2007}, which focused on adaptive momentum control during post-capture operations, our work emphasizes the accurate identification of unknown target inertia while ensuring the physical consistency of the estimated parameters. The proposed approach achieved physically consistent estimation results.}

As future work, we plan to validate the proposed method on a real robotic platform, such as the air-floating robot developed in our laboratory. The estimation performance will be quantitatively compared with existing approaches, such as the Extended Kalman Filter (EKF), to evaluate accuracy and robustness.

In addition, we aim to identify effective manipulator motions that enhance the accuracy of inertia parameter estimation 
{while maintaining controller stability, a problem commonly referred to as the exciting motion problem}. To achieve this, we plan to formulate a trajectory optimization problem by: 1) Parameterizing the joint trajectory using a truncated Fourier series to ensure smooth and periodic motion; 2) Optimizing the Fourier coefficients to maximize the quality of the regressor matrix, for example, by minimizing its condition number or maximizing the determinant of the associated information matrix
{, while incorporating stability-related constraints from the controller.}

Finally, we plan to apply this estimation framework to an adaptive control scheme, enabling online identification and real-time control updates during on-orbit manipulation. This integration is expected to improve tracking performance and robustness in dynamic space environments, which is particularly crucial for active debris removal missions, where an unknown target becomes part of the system dynamics.


\bibliography{./IEEEabrv,bibliography.bib}

\end{document}

%% file: macros.tex
\newcommand{\fig}[1]{Fig.~\ref{#1}}

\newcommand{\tab}[1]{Table~\ref{#1}}

\newcommand{\eq}[1]{(\ref{#1})}

\def\epsgaiji#1{\leavevmode\kern-0.025zw\raise-.37zh\hbox{%
  \epsfile{file=#1,width=1.05zw}}\kern-0.025zw}
\newcommand{\MARU}[1]{{\ooalign{\hfil#1\/\hfil\crcr\raise.167ex\hbox{\mathhexbox20D}}}}

\usepackage{array}


%% file: table/ground_logdet_hand.tex
\begin{tabular}{lcc}
\hline
& Error 0\% & Error 2.5\% \\ 
\hline \hline
Target \SI{5}{kg} &  0.021681 & 0.024287 \\ 
Target \SI{10}{kg} & 0.010557 & 0.047 \\ 
\hline 
\end{tabular}

%% file: table/orbital_logdet_hand.tex
\begin{tabular}{lrr}
\hline
& Force & Momentum \\ 
\hline \hline
 Target 5kg~ (error 2.5\% )& 0.31147 & 0.17938 \\ 
 Target 10kg (error 2.5\%) & 0.21583 & 0.090261 \\ 
 Target 50kg (error 0\%) & 0.13266 & 0.007674 \\ 
 Target 50kg (error 2.5\%) & 0.145173 & 0.01927 \\ 

\hline 
\end{tabular}